\documentclass[10pt,twocolumn,letterpaper]{article}

\usepackage{iccv}
\usepackage{times}
\usepackage{epsfig}
\usepackage{graphicx}
\usepackage{amsmath}
\usepackage{amssymb}
\usepackage{adjustbox}
\usepackage{array}
\usepackage{multirow}
\usepackage{comment}
\usepackage{authblk}
\newcommand*{\email}[1]{\texttt{#1}}
\newcommand*{\affaddr}[1]{#1} 

\usepackage[pagebackref=true,breaklinks=true,letterpaper=true,colorlinks,bookmarks=true]{hyperref}
\usepackage{cleveref}

\iccvfinalcopy 
\def\iccvPaperID{****} 

\begin{document}

\title{3C-Net: Category Count and Center Loss for Weakly-Supervised Action Localization}
\author{%
\vspace{-1cm}
Sanath Narayan, Hisham Cholakkal, Fahad Shahbaz Khan, Ling Shao \\ 
\vspace{-0.25cm}
\affaddr{Inception Institute of Artificial Intelligence, UAE}\\
\email{\small firstname.lastname@inceptioniai.org} \\
}
 
\maketitle

\begin{abstract}
   Temporal action localization is a challenging computer vision problem with numerous real-world applications. Most existing methods require laborious frame-level supervision to train action localization models. In this work, we propose a framework, called 3C-Net, which only requires video-level supervision (weak supervision) in the form of action category labels and the corresponding count. We introduce a novel formulation to learn discriminative action features with enhanced localization capabilities. Our joint formulation has three terms: a classification term to ensure the separability of learned action features, an adapted multi-label center loss term to enhance the action feature discriminability and a counting loss term to delineate adjacent action sequences, leading to improved localization. Comprehensive experiments are performed on two challenging benchmarks: THUMOS14 and ActivityNet 1.2. Our approach sets a new state-of-the-art for weakly-supervised temporal action localization on both datasets. On the THUMOS14 dataset, the proposed method achieves an absolute gain of 4.6\% in terms of mean average precision (mAP), compared to the state-of-the-art~\cite{wtalc}. Source code is available at \url{https://github.com/naraysa/3c-net}.
   \vspace{-0.2cm}
\end{abstract}

\begin{figure*}[t]
    \centering
    \includegraphics[width=1\linewidth,height=0.16\linewidth]{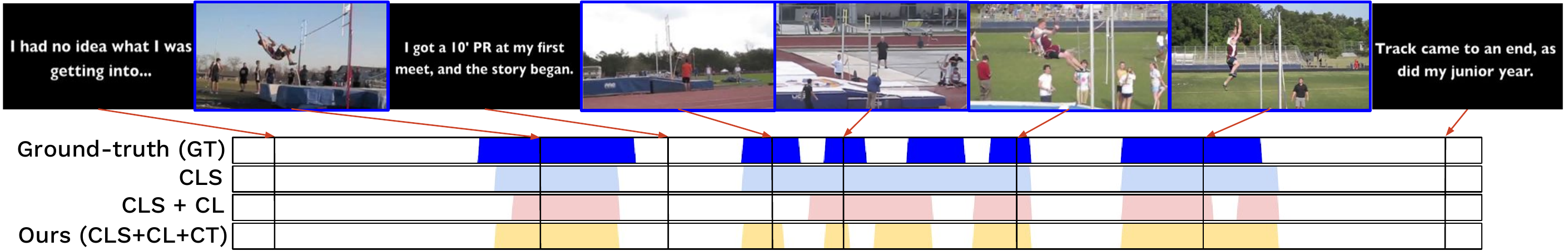}
    \caption{Predicted action proposals for a video clip containing \emph{PoleVault} action category from THUMOS14 dataset. Sample frames from the video are shown in the top row. Frames containing actions have a \emph{blue} border. GT indicates the ground-truth segments in the video containing the action. The network trained with classification loss term alone (CLS) inaccurately merges the four actions instances in the middle as a single instance. The network trained on classification and center loss terms (denoted as CLS + CL) improves the action localization but only partially delineates the merged action instances. The proposed 3C-Net framework, denoted as Ours (CLS + CL + CT), trained using a joint formulation of classification, center and counting loss terms, delineates the adjacent action instances in the middle. White regions in the timeline indicate background regions which do not contain actions of interest.}
    \label{fig_intro}\vspace{-0.22cm}
\end{figure*}

\section{Introduction}
Temporal action localization in untrimmed videos is a challenging problem due to intra-class variations, cluttered background, variations in video duration, and changes in viewpoints. In temporal action localization, the task is to find the start and end time (temporal boundaries or extent) of actions in a video. Most existing action localization approaches are based on strong supervision~\cite{fv-dtf,talnet,cdc,ssn,scnn,rc3d}, requiring manually annotated ground-truth temporal boundaries of actions during training. However, frame-level action boundary annotations are expensive compared to video-level action label annotations. Further, unlike object boundary annotations in images, manual annotations of temporal action boundaries are more subjective and prone to large variations~\cite{action-snippet,action-extent}.
Here, we focus on learning to temporally localize actions using only video-level supervision, commonly referred to as \emph{weakly-supervised} learning.

\vspace{-0.01cm}
Weakly-supervised temporal action localization has been investigated using different types of weak labels, \eg, action categories~\cite{hideseek,untrimnets,stpn}, movie scripts~\cite{TAL-movie-scripts,actor-action} and sparse spatio-temporal points~\cite{point-supervision}. Recently, Paul \etal ~\cite{wtalc} proposed an action localization approach, demonstrating state-of-the-art results, using video-level category labels as the weak supervision.  In their approach ~\cite{wtalc}, a formulation based on co-activity similarity loss is introduced which distinguishes similar and dissimilar temporal segments (regions) in paired videos containing same action categories. This leads to improved action localization results. However, the formulation in ~\cite{wtalc} puts a constraint on the mini-batch, used for training, to mostly contain paired videos with actions belonging to the same category. In this work, we look into an alternative formulation that allows the mini-batch to contain diverse action samples during training.

We propose a \textcolor{black}{framework, called 3C-Net, using a} novel formulation to learn discriminative action features with enhanced localization capabilities using video-level supervision. As in~\cite{stpn,wtalc}, our formulation contains a classification loss term that ensures the inter-class separability of learned features, for video-level action classification. However, this separability at the global video-level alone is insufficient for accurate action localization, which is generally a local temporal-context classification. This can be observed in Fig.~\ref{fig_intro}, where the network trained with classification loss alone, denoted as 'CLS', localizes multiple instances of an action (central portion of the timeline) as a single instance.
We therefore introduce two additional loss terms in our formulation that ensure both the discriminability of action categories at the global-level and separability of instances at the local-level.

\vspace{-0.01cm}
The first additional term in our formulation is the center loss~\cite{center_loss}, introduced here for multi-label action classification. Originally designed for the face recognition problem~\cite{center_loss}, the objective of the center loss term is to reduce the intra-class variations in the feature representation of the training samples. This is achieved by learning the class-specific centers and penalizing the distance between the features and their respective class centers. However, the standard center loss operates on training samples representing single-label instances. This prohibits its direct applicability in our multi-label action localization settings. We therefore propose to use a class-specific attention-based feature aggregation scheme to utilize multi-label action videos for training with center loss. As a result, a discriminative feature representation is obtained for improved localization. This improvement over 'CLS' can be observed in Fig.~\ref{fig_intro}, where the network trained using the classification and center loss terms, denoted as 'CLS + CL', partially solves the incorrect grouping of multiple action instances.

The final term in our formulation is a counting loss term, which enhances the separability of action instances at the local-level. Count information has been previously exploited in the image domain for object delineation~\cite{obj-c-wsl,cholakkal2019object}. In this work, the counting loss term incorporates information regarding the frequency of an action category in a video. The proposed loss term minimizes the distance between the predicted action count in a video and the ground-truth count. Consequently, the prediction scores sum up to a positive value within action instances and zero otherwise, leading to improved localization. This can be observed in Fig.~\ref{fig_intro}, where the proposed 3C-Net trained using all the three loss terms, denoted as 'Ours (CLS + CL + CT)', delineates all four adjacent action instances, thereby leading to improved localization. Our counting term utilizes video-level action count and does not require user-intensive action location information (e.g. temporal boundaries).

\subsection{Contributions} 
We introduce a weakly-supervised action localization framework, 3C-Net, with a novel formulation. Our formulation consists of a classification loss to ensure inter-class separability, a multi-label center loss to enhance the feature discriminability and a counting loss to improve the separability of adjacent action instances. The three loss terms in our formulation are jointly optimized in an end-to-end fashion. 
To the best of our knowledge, we are the first to propose a formulation containing center loss for multi-label action videos and counting loss to utilize video-level action count information for weakly-supervised action localization.

We perform comprehensive experiments on two benchmarks: THUMOS14~\cite{thumos14} and ActivityNet~1.2~\cite{activitynet}. Our joint formulation significantly improves the baseline containing only classification loss term. Further, our approach sets a new state-of-the-art on both datasets and achieves an absolute gain of 4.6\% in terms of mAP, compared to the best existing weakly-supervised method on THUMOS14. 

\section{Related Work}
Temporal action localization in untrimmed videos is a challenging problem that has gained significant attention in recent years. This is evident in popular challenges, such as THUMOS~\cite{thumos14} and ActivityNet~\cite{activitynet}, where a separate track is dedicated to the problem of temporal action localization in untrimmed videos. Weakly-supervised action localization mitigates the need for temporal action boundary annotations and is therefore an active research problem. In the standard settings, only action category labels are available to train a localization model. 
Existing approaches have investigated different weak supervision strategies for action localization. The work of~\cite{hideseek,stpn,untrimnets} use action category labels in videos for temporal localization, whereas~\cite{point-supervision} uses point-level supervision to spatio-temporally localize the actions. \cite{order-constraints-kuehne,order-constraints-bojan} exploit the order of actions in a video as a weak supervision cue. The work of~\cite{TAL-movie-scripts,auto-annotation}
use video subtitles and movie scripts to obtain coarse temporal localization for training, while~\cite{actor-action} utilizes actor-action pairs extracted from scripts for learning spatial actor-action localization. Recent work of~\cite{obj-c-wsl} shows that object counting with image-level supervision is less expensive, in terms of annotation cost, compared to instance-level supervision (\eg, bounding-box). In this work, we propose to use action instance count as an additional cue for weakly-supervised action localization.

State-of-the-art weakly-supervised action localization methods utilize both appearance and motion features, typically extracted from backbone networks trained for the action recognition task. The work of \cite{untrimnets} proposes a framework that consists of a classification and a selection module for classifying the actions and detecting the relevant temporal segments, respectively. The approach uses a two-stream Temporal Segment Network~\cite{tsn} as its backbone and employs a classification loss for training. In~\cite{stpn}, a two-stream architecture is used to learn temporal class activation maps and a class-agnostic temporal attention. Their combination is then used to localize the human actions. Classification and sparsity-based losses are used to learn the activation maps and temporal attention, respectively. Recently,~\cite{wtalc} proposed a framework to learn temporal localization from video-level labels, where a classification loss and a triplet loss for matching similar segments of an action category in paired videos is employed. In this work, we propose a joint formulation with explicit loss terms to ensure the separability of learned action features, enhance the feature discriminability and delineate adjacent action instances.

\section{Method}
In this section, we first describe the feature extraction scheme used in our approach. We then present our overall architecture followed by a detailed description of the different loss terms in the proposed formulation.

\noindent\textbf{Feature Extraction}: As in~\cite{stpn,wtalc}, we use Inflated 3D (I3D) features extracted from the RGB and flow I3D deep networks~\cite{kinetics}, trained on the Kinetics dataset, to encode appearance and motion information, respectively.
A video is divided into non-overlapping segments, each consisting of 16 frames. The input to the RGB and flow I3D networks are the color
and the corresponding optical flow frames of a segment, respectively.
A $D$-dimensional output I3D feature per segment, from each of the two networks, is used as input to the respective RGB and flow streams in our architecture.
\vspace{-0.45cm}
\begin{figure*}[t]
    \centering
    \includegraphics[width=0.97\textwidth,height=0.393\textwidth]{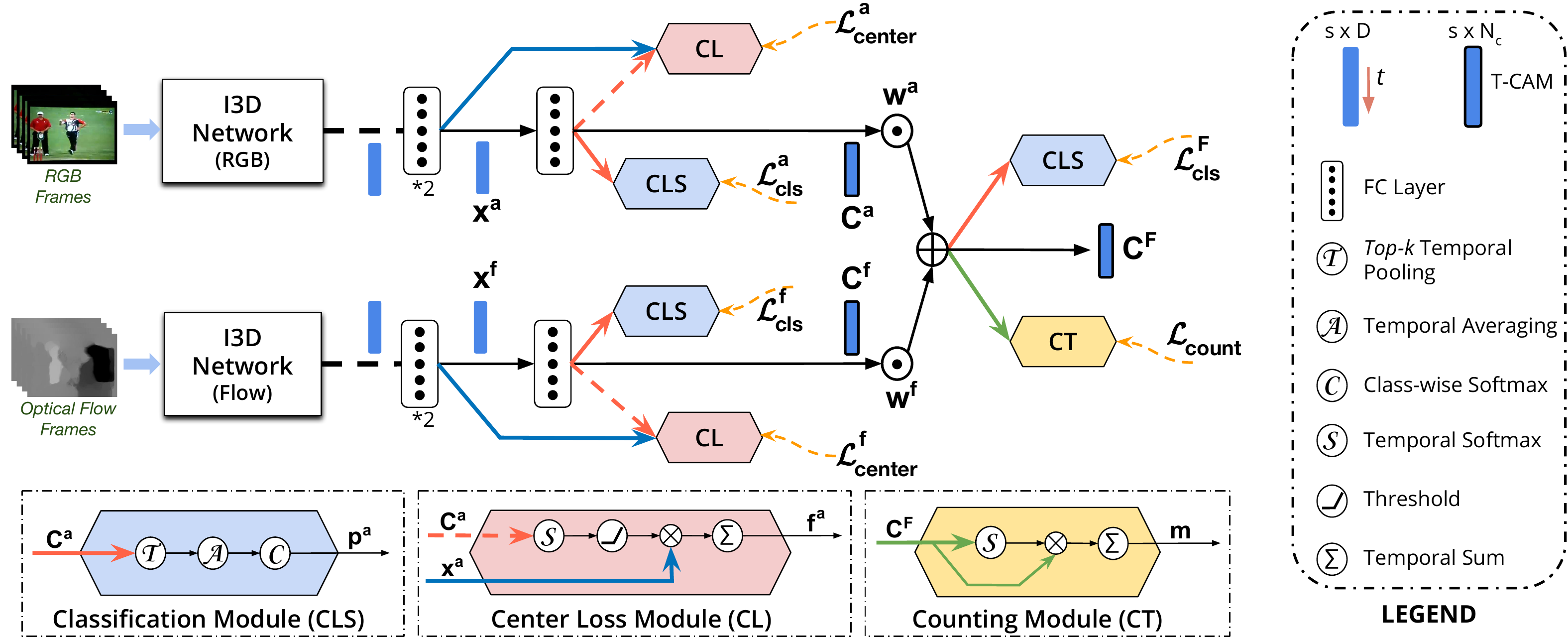}
    \vspace{-0.05cm}
    \caption{Our overall architecture (3C-Net) with different loss terms (classification, center and counting), and the associated modules.
    The architecture is based on a two-stream model (RGB and flow) with an associated backbone feature extractor in each stream. Both streams are structurally identical and consist of two fully-connected layers (FC). The outputs of the final FC layer in both streams are the temporal class activation maps (T-CAM), $\mathbf{C}^a$ for RGB and $\mathbf{C}^f$ for flow.
    The two T-CAMs are weighted by class-specific parameters ($\mathbf{w^a}$ and $\mathbf{w^f}$) and combined in a late fusion manner. The resulting T-CAM, $\mathbf{C}^F$, is used for inference. The modules for the different loss terms do not have learnable parameters and are shown separately in the bottom row with sample inputs and corresponding outputs for clarity.
    Both center ($\mathcal{L}_{center}^a$, $\mathcal{L}_{center}^f$) and classification ($\mathcal{L}_{cls}^a$, $\mathcal{L}_{cls}^f$) losses are applied to each of the two streams ($\mathbf{C}^a$ and $\mathbf{C}^f$) whereas the classification ($\mathcal{L}_{cls}^F$) and counting ($\mathcal{L}_{count}$) loss are applied to the fused representation ($\mathbf{C}^F$). Superscripts $a$, $f$ and $F$ denote appearance (RGB), flow and final, respectively.
    Color-coded arrows denote the association between the features in the network and the respective modules. 
    }
    \label{fig_overall_arch}\vspace{-0.2cm}
\end{figure*}

\subsection{Overall Architecture\label{sec_arch}}
Our overall 3C-Net architecture is shown in Fig.~\ref{fig_overall_arch}. In our approach, both appearance (RGB) and motion (flow) features are processed in parallel streams. The two streams are then fused at a later stage of the network. Both streams are structurally identical in design. Each stream in our network comprises of three fully-connected (FC) layers. Guided by the center loss~\cite{center_loss}, the first two FC layers learn to transform the I3D features into a discriminative
intermediate feature representation. The final FC layer projects the intermediate features into the action category space under the guidance of the classification loss. The outputs of the final FC layer represent the sequence of classification scores for each action over time. This class-specific 1D representation, similar to the 2D class activation map in object detection~\cite{obj-det-cam}, is called temporal class activation map (T-CAM), as in~\cite{stpn}.

Given a training video $v_i$, let $\mathbf{y}_i \in \mathbb{R}^{N_c}$ denote the ground-truth multi-hot vector indicating the presence or absence of an action category in $v_i$, where $i \in [1,N]$. Here, $N$ is the number of videos and $N_c$ is the number of action classes in the dataset. Let $\mathbf{x}_i^a$, $\mathbf{x}_i^f$ $\in \mathbb{R}^{s_i\times D}$ denote the intermediate features (outputs of the second FC layer) in the two streams, respectively. Here, $s_i$ denotes the length (number of segments) of the video $v_i$.  
The output of the final FC layers represent the T-CAMs, denoted by $\mathbf{C}_i^a$, $\mathbf{C}_i^f \in \mathbb{R}^{s_i \times N_c}$, for the RGB and flow streams, respectively. 
The two T-CAMs ($\mathbf{C}_i^a$ and $\mathbf{C}_i^f$) are weighted by learned class-specific parameters, $\mathbf{w}^a, \mathbf{w}^f \in \mathbb{R}^{N_c}$, and later combined by addition to result in the final T-CAM, $\mathbf{C}_i^F \in \mathbb{R}^{s_i \times N_c}$. The learning of the final T-CAM, $\mathbf{C}_i^F$ is guided by the classification and counting loss terms. 
Consequently, our 3C-Net framework is trained using the overall loss formulation, 
\vspace{-0.1cm}
\begin{equation}
    \label{eqn_total_loss}
    \mathcal{L} = \mathcal{L}_{cls}  + \alpha \mathcal{L}_{center} + \beta \mathcal{L}_{count}
    \vspace{-0.1cm}
\end{equation}
\noindent where $\mathcal{L}_{cls}$, $\mathcal{L}_{center}$ and $\mathcal{L}_{count}$ denote the classification loss, center loss and counting loss terms, respectively. The respective weights for the center loss and counting loss terms are denoted by $\alpha$ and  $\beta$. Next, we describe the three loss terms utilized in the proposed formulation.

\subsection{Classification Loss\label{sec_mill}}
The classification loss term is used in our formulation to ensure the inter-class separability of the features at the video-level and tackles the problem of multi-label action classification in the video. We utilize the cross-entropy classification loss as in~\cite{untrimnets,wtalc}, to recognize different action categories in a video. The number of segments per video varies greatly in untrimmed videos. Hence, the \emph{top-k} values per category (where $k=\lceil s_i/8\rceil$, is proportional to the length, $s_i$, of the video) of a T-CAM\footnote{\label{footnote_loss_for_single_stream}For brevity, the loss computation is explained in detail for the RGB stream using the superscript $a$ (denoting appearance) for the variables.} ($\mathbf{C}_i^a$) are selected, as in~\cite{wtalc}. This results in a representation of size $k\times N_c$, for the video. Further, a temporal averaging is performed on this representation to obtain a class-specific encoding, $\mathbf{r}_i^a \in \mathbb{R}^{N_c}$, for the T-CAM, $\mathbf{C}_i^a$. Consequently, a probability mass function (pmf), $\mathbf{p}_i^a \in \mathbb{R}^{N_c}$, is computed using
\vspace{-0.1cm}
\begin{equation}
    \label{eqn_cls_pred}
    \mathbf{p}_i^a(j) = \frac{\exp({\mathbf{r}_i^a(j)})}{\sum_l\exp({\mathbf{r}_i^a(l)})}
\end{equation}
where $j \in [1,N_c]$ denotes the action category. As shown in Fig.~\ref{fig_overall_arch}, the 'Classification Module (CLS)' performs \emph{top-k} temporal pooling, averaging and category-wise softmax operations and outputs a predicted pmf, $\mathbf{p}_i^a$ for an input, $\mathbf{C}_i^a$. The multi-hot encoded ground-truth action labels $\mathbf{y}_i$, are $l_1$-normalized to generate a ground-truth pmf, $\mathbf{q}_i$. The classification loss is then represented as the cross-entropy between $\mathbf{p}_i^a$ and $\mathbf{q}_i$. Let $\mathcal{L}_{cls}^a = -\mathbb{E}[\mathbf{q}_i^Tlog(\mathbf{p}_i^a)]$ denote the classification loss for the RGB stream, where $\mathbf{p}_i^a$ is the pmf computed from $\mathbf{C}_i^a$. 
The loss for the flow stream T-CAM $\mathbf{C}_i^f$ and the final T-CAM $\mathbf{C}_i^F$, are computed in a similar manner. The total classification loss, $\mathcal{L}_{cls}$, is then given by,

\vspace{-0.1cm}
\begin{equation}
    \label{eqn_cls_loss}
    \mathcal{L}_{cls} = \mathcal{L}_{cls}^a + \mathcal{L}_{cls}^f + \mathcal{L}_{cls}^F
\end{equation}

\subsection{Center Loss for Multi-label Classification\label{sec_center_loss}}
We adapt and integrate the center loss term~\cite{center_loss} in our overall formulation to cluster the features of different categories such that the same action category features are grouped together.
The center loss learns the cluster centers of each action class and penalizes the distance between the features and the corresponding class centers. The objective of the classification loss, commonly employed in action localization, is to ensure the inter-class separability of learned features, whereas the center loss aims to enhance their discrminability through action-specific clustering and minimizing the intra-class variations. However, the standard center loss, originally proposed for face recognition~\cite{center_loss}, operates on training samples representing single-label instances. This hinders its usage in multi-label weakly-supervised action localization settings, where training samples (videos) contain multiple action categories. To counter this issue, we employ an attention-based per-class feature aggregation strategy to utilize videos with multiple action categories for training with the center loss. To the best of our knowledge, we are the first to introduce the center loss with multi-label training samples for weakly supervised action localization.

In the proposed 3C-Net framework, the center loss is applied on the features\textsuperscript{\ref{footnote_loss_for_single_stream}},
$\mathbf{x}_i^a$ (output of the penultimate FC layer as in Fig.~\ref{fig_overall_arch}). 
Typically, videos vary in length ($s_i$) and contain multiple action classes. 
Additionally, the action duration may be relatively short in untrimmed videos.
Hence, aggregating category-specific features by considering only the high attention regions of those categories in the video is required.
We perform the feature aggregation step on $\mathbf{x}_i^a$ and compute a single feature $\mathbf{f}_i^a(j) \in \mathbb{R}^D$ if $\mathbf{y}_i(j) \neq 0$ (\ie, if the action category $j$ is present in video $v_i$). In the case of action categories which are not present in a video, the feature aggregation step is not performed, since these categories will not have a meaningful feature representation in that video.
To this end, we first compute the attention, $\mathbf{a}_i^a \in \mathbb{R}^{s_i \times N_c}$, over time $t$, for a category $j$, using
\begin{equation}
    \label{eqn_temporal_atn}
    \mathbf{a}_i^a(t,j) = \frac{\exp(\mathbf{C}_i^a(t,j))}{\sum_{l} \exp(\mathbf{C}_i^a(l,j))}
\end{equation}
where $\mathbf{C}_i^a$ represents the RGB stream T-CAM for video $v_i$. A threshold, 
$\tau_j=$median$(\mathbf{a}_i^a(j))$
is used to set the attention weights less than $\tau_j$ to 0 (\ie $\mathbf{a}_i^a(t,j) = 0$, if $\mathbf{a}_i^a(t,j) <\tau_j)$). Here, $s_i$ is the length of the video. This thresholding enables feature aggregation from category-specific high-attention regions of the video. The resulting aggregated features, $\mathbf{f}_i^a(j)$, are then used with the center loss. 
The aggregated feature $\mathbf{f}_i^a(j)$ is computed using 
\begin{equation}
    \label{eqn_temporal_feat_aggr}
    \mathbf{f}_i^a(j) = \frac{\textstyle\sum_t \mathbf{a}_i^a(t,j)\mathbf{x}_i^a(t)}{\textstyle\sum_t \mathbf{a}_i^a(t,j)}
\end{equation}
As shown in Fig.~\ref{fig_overall_arch}, the 'Center Loss Module (CL)' implements Eq.~\ref{eqn_temporal_atn}~and~\ref{eqn_temporal_feat_aggr} for each stream, using the outputs of the FC layers of the respective stream. 
Let $\mathbf{c}_j^a \in \mathbb{R}^D$ be the cluster center associated with the action category $j$. Following~\cite{center_loss}, the center loss and the update for center $\mathbf{c}_j^a$, used in our multi-label formulation, are given by, 
\begin{equation}
    \mathcal{L}_{center}^a = \frac{1}{N}{\textstyle\sum\limits_i}{\textstyle\sum\limits_{j:\mathbf{y}_i(j)=1}}||\mathbf{f}_i^a(j) - \mathbf{c}_j^a||^2_2
\end{equation}
\begin{equation}
    \Delta \mathbf{c}_j^a = \frac{\textstyle\sum_{i:\mathbf{y}_i(j)=1}(\mathbf{c}_j^a-\mathbf{f}_i^a(j))}{1+\textstyle\sum_i \mathbf{y}_i(j)}
\end{equation}
For every category $j$, present in a mini-batch, the corresponding center, $\mathbf{c}_j^a$ is updated using its $\Delta \mathbf{c}_j^a$ during training.
The loss for the flow stream, $\mathcal{L}_{center}^f$, is also computed in a similar manner. The total center loss is then given by,
\begin{equation}
    \label{eqn_center_loss}
    \mathcal{L}_{center} = \mathcal{L}_{center}^a + \mathcal{L}_{center}^f 
\end{equation}

\subsection{Counting Loss\label{sec_count_loss}}
In this work, we propose to use auxiliary count information in addition to standard action category labels for weakly-supervised action localization. Here, count refers to the number of instances of an action category occurring in a video. As discussed earlier, integrating count information enhances the feature representation and delineation of temporally adjacent action instances in the video, leading to an improved temporal localization. In our 3C-Net framework, the counting loss is applied on the final T-CAM, $\mathbf{C}_i^F$.

To compute the predicted count, first, the element-wise product of the category-specific temporal attention and the final T-CAM, $\mathbf{C}_i^F$, is performed. The resulting attention-weighted T-CAM is equivalent to a density map \cite{cholakkal2019object} of the action category, and its summation yields the predicted count of that category.
Let the attention for action category $j$ be $\mathbf{a}_i^F(j)$, which is computed using the final T-CAM, similar to 
Eq.~\ref{eqn_temporal_atn}. The predicted count for category $j$ is given by,  
\begin{equation}
\label{eqn_pred_count}
\mathbf{m}_i(j) = \sum_{t} \mathbf{a}_i^F(t,j) \mathbf{C}_i^F(t,j)
\vspace{-0.2cm}
\end{equation}
\noindent where $\mathbf{m}_i(j)$ represents the sum of activation weighted by the temporal attention, over time for the $j^{th}$ action category. As shown in Fig.~\ref{fig_overall_arch}, the 'Counting Module (CT)' implements Eq.~\ref{eqn_temporal_atn}~and~\ref{eqn_pred_count} for the final T-CAM, $\mathbf{C}_i^F$. Temporal attention weighting ignores the background video segments not containing the action category $j$.
 
In the context of action localization, we observe that videos with a higher action count tend to have higher errors in count prediction during training. 
Training with absolute error results in an inferior T-CAM, since the mini-batch loss will be dominated by the count prediction error for the videos with a higher action count.
To tackle this issue, we use a simple yet effective weighting strategy, where errors are inversely weighted depending on the action count in a video. A lower weight is assigned when the action count in a video is high and vice versa. 
The weighting penalizes the count error (\emph{ce}) more at lower ground-truth count (GTC) compared to the same magnitude of \emph{ce} at higher GTC. \Eg, \emph{ce} = 1 at GTC of 5 is emphasized over \emph{ce} = 1 at GTC of 100.
To obtain a relative error for per-category count prediction, we divide the absolute error by the GTC of the categories present in the video. Absolute error is used for the action categories that are not present in a video to ensure that their predicted count is zero. The counting loss is then given by,
\begin{equation}
    \mathcal{L}_{count}^+ = \frac{1}{N}{\textstyle\sum\limits_i}{\textstyle\sum\limits_{j:\mathbf{n}_i(j)> 0}}\frac{|\mathbf{m}_i(j) - \mathbf{n}_i(j)|}{\mathbf{n}_i(j)} \nonumber
\end{equation}
\begin{equation}    
    \mathcal{L}_{count}^- = \frac{1}{N}{\textstyle\sum\limits_i}{\textstyle\sum\limits_{j:\mathbf{n}_i(j)=0}}|\mathbf{m}_i(j)| \nonumber
\end{equation}
\begin{equation}  
   \label{eqn_count_loss}
    \mathcal{L}_{count} = \mathcal{L}_{count}^+ + \lambda\mathcal{L}_{count}^-
\end{equation}
where $\mathbf{n}_i \in \mathbb{R}^{N_c}$ is the ground-truth count label and $\lambda$ is a hyper-parameter, typically set to $10^{-3}$ to compensate for the ratio of positive to negative instances for an action class.

To summarize, the loss terms in our overall formulation enhance the separability and  discriminability of the learned features and improve the delineation of adjacent action instances. Consequently, a disrcriminative and improved T-CAM representation is obtained.

\subsection{Classification and Localization using T-CAM\label{sec_cls_loc}}
After training the 3C-Net, the CLS module (see Fig.~\ref{fig_overall_arch} and Eq.~\ref{eqn_cls_pred}) is used to compute the action-class scores (pmf) at the video-level using the final T-CAM, for the action classification task. Similar to~\cite{untrimnets,wtalc}, we use the computed pmf without threshold, for evaluation. For the action localization task, detections are obtained using a similar approach used in~\cite{wtalc}. Detections in a video are generated for the action categories with average \emph{top-k} score above 0 (\ie for categories in set \{$j:\mathbf{r}_i^F(j)>0$\}, where $\mathbf{r}_i^F$ is computed as in Sec.~\ref{sec_mill} using the final T-CAM). For a category $j$ in the obtained set, continuous video segments between successive time instants when T-CAM goes above and below threshold $\eta$, correspond to a valid action detection. The resulting detections of an action category are non-overlapping. A weighted sum of the highest T-CAM value with in the detection and the category score for the video, corresponds to the score of a detection. The detection with the highest score that is overlapping (above IoU threshold) with the ground-truth is considered true-positive during evaluation.

\section{Experiments \label{sec_exp_setup_results}}

\subsection{Experimental Setup\label{sec_exp_setup}}
\noindent\textbf{Datasets}: The proposed 3C-Net is evaluated for temporal action localization on two challenging datasets containing untrimmed videos with varying degree of activity duration. 

\noindent\textbf{THUMOS14}~\cite{thumos14} dataset contains 1010 validation and 1574 test videos from 101 action categories. Out of these, 20 categories have temporal annotations in 200 validation and 213 test videos. The dataset is challenging, as it contains an average of 15 activity instances per video. Similar to ~\cite{stpn,wtalc}, we use the validation set for training and test set for evaluating our framework. 
\noindent\textbf{ActivityNet 1.2}~\cite{activitynet} dataset has 4819 training, 2383 validation and 2480 testing videos from 100 activity categories. Note that the test set annotations for this dataset are withheld. There are an average of 1.5 activity instances per video. As in ~\cite{autoloc,wtalc}, we use the training set to train and validation set to test our approach. 

\noindent\textbf{Count Labels}: The ground-truth count labels for the videos in both datasets are generated using the available temporal action segments information. The total number of segments of an action category in a video is the ground-truth count video-label for the respective category. This was done to use the available annotations and avoid re-annotations. However, for a new dataset, action count can be independently annotated, without requiring action segment information.

\noindent\textbf{Evaluation Metric}: We follow the standard protocol, provided with the two datasets, for evaluation. The evaluation protocol is based on mean Average Precision (mAP) for different intersection over union (IoU) values for the action localization task. For the multi-label action classification task, we use the mAP computed from the predicted video-level scores for evaluation.

\noindent\textbf{Implementation Details}:
We use an alternate mini-batch training approach to train the proposed 3C-Net framework. Since, the count labels are available at the video-level, all the segments of a video are required for count prediction.  We use random temporal cropping of videos in alternate mini-batches to improve the generalization. Thus, the classification and center losses are used for every mini-batch training and the counting loss is applied \emph{only} on the alternate mini-batches containing the full-length video features.

In our framework, a TV-L1 optical flow~\cite{tvl1-flow} is used to generate the optical flow frames of the video. The I3D features of size $D=1024$ per segment of 16 video frames are obtained after spatio-temporal average pooling of \emph{Mixed\_5c} layers from the RGB and Flow I3D networks. These I3D features are then used as input to our framework. As in ~\cite{stpn,wtalc}, the backbone networks are not finetuned. Our 3C-Net is trained with a mini-batch size of 32 using the Adam~\cite{adam} optimizer with $10^{-4}$ learning rate and 0.005 weight decay. \textcolor{black}{The centers $\mathbf{c}_j$ are learned using the SGD optimizer with 0.1 learning rate.} For both datasets, we set $\alpha$ in Eq.~\ref{eqn_total_loss} to 10${}^{-3}$ since the center loss penalty is a squared error loss with a higher magnitude compared to other loss terms. We set $\beta$ in Eq.~\ref{eqn_total_loss} to 1 and 0.1 for the THUMOS14 and ActivityNet 1.2 datasets, respectively. 
$\eta$ is set to $0.5[\min(\mathbf{C}_i^F(j))+\max(\mathbf{C}_i^F(j))]$ for a $j^{th}$ category T-CAM in THUMOS14. \textcolor{black}{Due to the nature of actions in ActivityNet 1.2, W-TALC~\cite{wtalc} approach uses the Savitzky-Golay filter~\cite{savitzky1964smoothing} for post-processing the T-CAMs. Here, we use a learnable temporal convolution filtering (kernel size=$13$, dilation=$2$) and set $\eta$ to 0.}

\begin{table}[t]
\centering
\adjustbox{width=\linewidth}{
\begin{tabular}{|l|c|c|c|c|c|c|}
\hline
\multicolumn{1}{|c|}{\multirow{2}{*}{\textbf{Approach}}} & \multicolumn{6}{c|}{\textbf{mAP @ IoU}} \\ \cline{2-7} 
\multicolumn{1}{|c|}{} & \textbf{0.1} & \textbf{0.2} & \textbf{0.3} & \textbf{0.4} & \textbf{0.5} & \textbf{0.7}  \\ 
 \hline \hline
FV-DTF~\cite{fv-dtf}${}^{+}$ & 36.6 & 33.6 & 27.0 & 20.8 & 14.4 & - \\ \hline
S-CNN~\cite{scnn}${}^{+}$ & 47.7 & 43.5 & 36.3 & 28.7 & 19.0 & 5.3 \\ \hline
CDC~\cite{cdc}${}^{+}$ & - & - & 40.1 & 29.4 & 23.3 & 7.9 \\ \hline
R-C3D~\cite{rc3d}${}^{+}$ & 54.5 & 51.5 & 44.8 & 35.6 & 28.9 & - \\ \hline
TAL-Net~\cite{talnet}${}^{+}$ & 59.8 & 57.1 & 53.2 & 48.5 & 42.8 & 20.8 \\ 
\hline \hline
UntrimmedNets~\cite{untrimnets} & 44.4 & 37.7 & 28.2 & 21.1 & 16.2 & 5.1 \\ \hline
STPN~\cite{stpn} & 52.0 & 44.7 & 35.5 & 25.8 & 16.9 & 4.3 \\ \hline
Autoloc~\cite{autoloc} & - & - & 35.8 & 29.0 & 21.2 & 5.8 \\ \hline
W-TALC~\cite{wtalc} & 53.7 & 48.5 & 39.2 & 29.9 & 22.0 & 7.3 \\
 \hline \hline
\textbf{Ours:} CLS + CL & 56.8 & 49.8 & 40.9 & 32.3 & 24.6 & 7.7 \\ \hline
\textbf{Ours: 3C-Net} & \textbf{59.1} & \textbf{53.5} & \textbf{44.2} & \textbf{34.1} & \textbf{26.6} & \textbf{8.1} \\ \hline
\end{tabular}
}
\vspace{-0.12cm}
\caption{\label{tab_th14_loc_sota}Action localization performance comparison (mAP) of our 3C-Net with state-of-the-art methods on THUMOS14 dataset. Superscript '$+$' for a method denotes that strong supervision is required for training. Our 3C-Net outperforms existing weakly-supervised methods and achieves an absolute gain of 4.6\%, at IoU=$0.5$, compared to the best weakly-supervised result~\cite{wtalc}.}\vspace{-0.3cm}
\end{table}

\subsection{State-of-the-art comparison\label{sec_exp_results}}
\noindent\textbf{Temporal Action Localization}: 
Tab.~\ref{tab_th14_loc_sota} shows the comparison of our 3C-Net method with existing approaches in literature on the THUMOS14 dataset. Superscript '$+$' for a method in Tab.~\ref{tab_th14_loc_sota} denotes that frame-level labels (strong supervision) are required for training. Our approach is denoted as '3C-Net'. We report mAP scores at different IoU thresholds. Both UntrimmedNets~\cite{untrimnets} and Autoloc~\cite{autoloc} use TSN~\cite{tsn} as the backbone, whereas STPN~\cite{stpn} and W-TALC~\cite{wtalc} use I3D networks similar to our framework. The STPN approach obtains an mAP of $16.9$ at IoU=$0.5$, while W-TALC achieves an mAP of $22.0$. 
Our approach CLS + CL, without any count supervision, outperforms all existing weakly-supervised action localization approaches. With the integration of count supervision, our 
3C-Net achieves an absolute gain of 4.6\%, in terms of mAP at IoU=$0.5$, over W-TALC~\cite{wtalc}. Further, a consistent improvement in performance is also obtained at other IoU thresholds.

Tab.~\ref{tab_actnet_loc_sota} shows the state-of-the-art comparison on the ActivityNet~1.2 dataset. We follow the standard evaluation protocol~\cite{activitynet} by reporting the mean mAP scores at different thresholds (0.5:0.05:0.95). 
Among the existing methods, the SSN approach~\cite{ssn} relies on frame-level annotations (strong supervision, denoted by superscript '$+$' in Tab.~\ref{tab_actnet_loc_sota}) for training and achieves a mean mAP score of 26.6.
\textcolor{black}{Our baseline approach, trained with the classification loss alone, achieves a mean mAP of 18.2. With only the center loss adaption, our approach achieves a mean mAP of 21.1 and surpasses all existing weakly-supervised methods. With the integration of count supervision, the performance further improves to 21.7 and outperforms the state-of-the-art weakly-supervised approach~\cite{wtalc} by 3.7\%, in terms of mean mAP. The relatively lower margin of improvement using count labels, compared to THUMOS14, is likely due to fewer multi-instance videos in training and noisy annotations in this dataset.}

\begin{table}[t]
\centering
\small
\begin{tabular}{|l|c|c|c|c|}
\hline
\multicolumn{1}{|c|}{\multirow{2}{*}{\textbf{Approach}}} & \multicolumn{4}{c|}{\textbf{mAP @ IoU}} \\ \cline{2-5} 
\multicolumn{1}{|c|}{} & \textbf{0.5} & \textbf{0.7} & \textbf{0.9} & \textbf{Avg*} \\ \hline
SSN~\cite{ssn}${}^{+}$ & 41.3 & 30.4 & 13.2 & 26.6 \\ \hline \hline
UntrimmedNets~\cite{untrimnets} & 7.4 & 3.9 & 1.2 & 3.6 \\ \hline
Autoloc~\cite{autoloc} & 27.3 & 17.5 & 6.8 & 16.0 \\ \hline
W-TALC~\cite{wtalc} & 37.0 & 14.6 & - & 18.0 \\ 
\hline \hline
\textbf{Ours:} CLS + CL & 35.4 & 22.9 & 8.5 & 21.1 \\ \hline
\textbf{Ours: 3C-Net} & \textbf{37.2} & \textbf{23.7} & \textbf{9.2} & \textbf{21.7} \\ \hline
\end{tabular}
\vspace{0.05cm}
\caption{\label{tab_actnet_loc_sota}Action localization performance comparison (mean mAP) of our 3C-Net with state-of-the-art methods on the ActivityNet 1.2 dataset. The mean mAP is denoted by Avg${}^*$. Note that SSN ~\cite{ssn} requires frame-level labels (strong supervision) for training. Our 3C-Net outperforms all existing weakly-supervised methods and obtains an absolute gain of 3.7\% in terms of mean mAP, compared to the state-of-the-art weakly-supervised W-TALC ~\cite{wtalc}.}\vspace{-0.3cm}
\end{table}

\noindent\textbf{Action Classification}:
We also evaluate our method for action classification. Tab.~\ref{tab_cls_sota} shows the comparison on  THUMOS14 and ActivityNet 1.2 datasets. 
\textcolor{black}{Our 3C-Net achieves a superior classification performance of 86.9, in terms of mAP, compared to existing methods on the THUMOS14 dataset and is comparable to W-TALC on ActivityNet 1.2.}

\begin{table}[t]
\centering
\adjustbox{width=0.9\linewidth}{   
\small
\begin{tabular}{|l|c|c|}
\hline
\multicolumn{1}{|c|}{\textbf{Approach}} & \textbf{THUMOS14} & \textbf{ActivityNet 1.2} \\ \hline
iDT+FV~\cite{idt-fv} & 63.1 & 66.5 \\ \hline
Objects + Motion~\cite{objects15000} & 71.6 & - \\ \hline
Two Stream~\cite{zisserman-two-stream} & 66.1 & 71.9 \\ \hline
C3D~\cite{c3d} & - & 74.1 \\ \hline
TSN~\cite{tsn} & 67.7 & 88.8 \\ \hline
UntrimmedNets~\cite{untrimnets} & 82.2 & 87.7 \\ \hline
W-TALC~\cite{wtalc} & 85.6 & \textbf{93.2} \\ 
\hline \hline
\textbf{Ours: 3C-Net} & \textbf{86.9} & 92.4 \\ \hline
\end{tabular}
}
 \vspace{0.05cm}
\caption{\label{tab_cls_sota}Action classification performance comparison (mAP) of our 3C-Net with state-of-the-art methods on the THUMOS14 and ActivityNet 1.2 datasets. On THUMOS14, our 3C-Net achieves superior classification result, compared to existing methods.}\vspace{-0.4cm}
\end{table}

\subsection{Baseline Comparison and Ablation Study\label{sec_ablation}}
\noindent\textbf{Baseline comparison}: 
Tab.~\ref{tab_baseline} shows the action localization performance comparison on THUMOS14 (at IoU=0.5). We also show the impact
of progressively integrating one contribution at a time in our 3C-Net framework. The baseline (CLS) trained using classification loss alone obtains a mAP score of 19.1. The integration of our multi-label center loss term (CLS + CL) significantly improves the performance by obtaining a mAP score of 24.6. The action localization performance is further improved to 26.6 mAP, by the integration of our counting loss term (CLS + CL + CT).

\begin{figure}[t]
    \centering
    \includegraphics[width=\linewidth,height=0.47\linewidth]{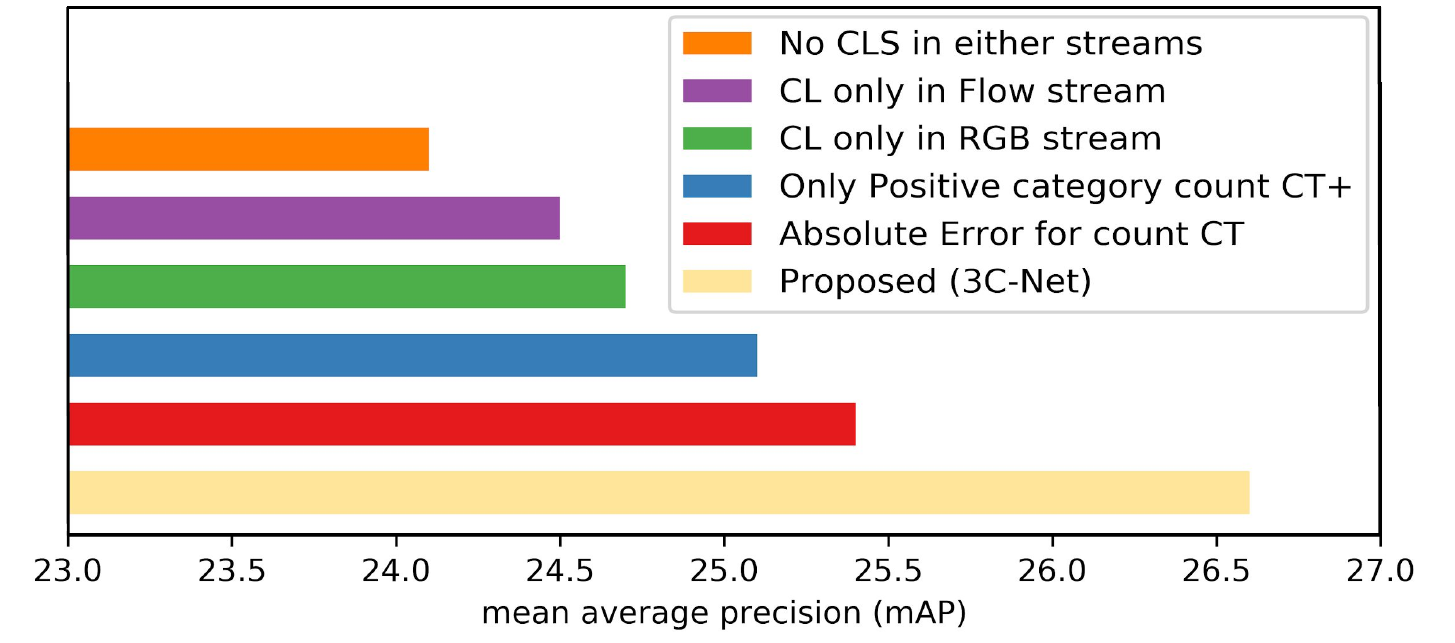}\vspace{-0.1cm}
    \caption{Ablation study with respect to difference design choices and different loss terms in our action localization framework on the THUMOS14 dataset. See text for details.}\vspace{-0.3cm}
    \label{fig_ablation}
\end{figure}

\noindent\textbf{Ablation study}: Fig.~\ref{fig_ablation} shows the results with respect to different design choices and impact of different loss terms in our action localization framework on the THUMOS14 dataset. All the experiments are conducted independently and show the deviation in performance relative to the proposed 3C-Net framework. The localization performance of our final proposed 3C-Net framework is shown as yellow bar. First, we show the impact of removing the classification loss in both the streams and retaining it only for the final T-CAM ($\mathbf{C}_i^F$). This results (orange bar) in a drop of 2.5\% mAP. Next, we observe that retaining center loss term only in the flow stream results in a drop of 2.1\% mAP (purple bar). Retaining the center loss term only in the RGB stream results in a drop of 1.9\% mAP (green bar). Afterwards, we observe that removing the negative category counting loss in Eq.~\ref{eqn_count_loss} results in a drop of 1.5\% mAP (blue bar). Further, replacing the relative error for counting loss with absolute error deteriorates the results by 1.2\% mAP (red bar). 
These results show that both our design choices and different loss terms contribute in the overall performance of our approach.

\begin{table}[t]
\centering
\adjustbox{width=0.95\linewidth}{
\small
 \begin{tabular}{|c|c|c|}
 \hline
 \textbf{Baseline: CLS} & \textbf{CLS + CL} & \textbf{3C-Net: CLS + CL + CT} \\
 \hline
 19.1 & 24.6 & \textbf{26.6} \\ \hline
 \end{tabular}
}
 \caption{\label{tab_baseline}Baseline action localization performance comparison (mAP) on THUMOS14 at IoU=0.5. Our 3C-Net achieves an absolute gain of 7.5\% in terms of mAP, compared to the baseline. }\vspace{-0.3cm}
\end{table}

\begin{figure*}[t]
    \centering
    \includegraphics[width=1\linewidth,height=0.15\linewidth]{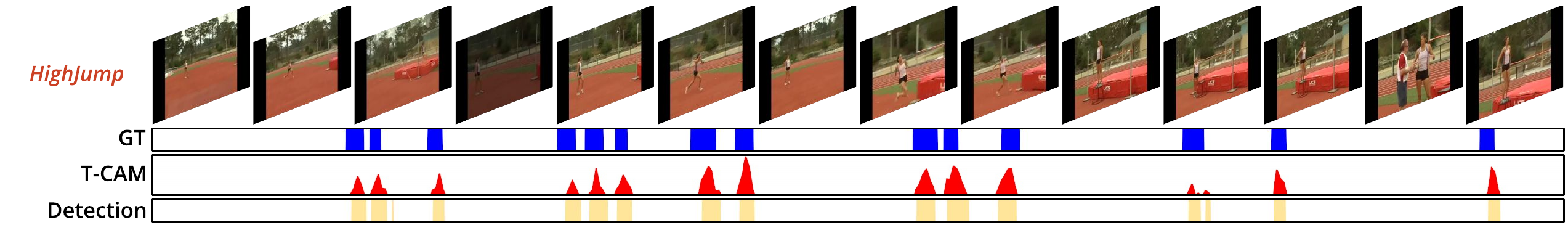} \\
    \includegraphics[width=1\linewidth,height=0.22\linewidth]{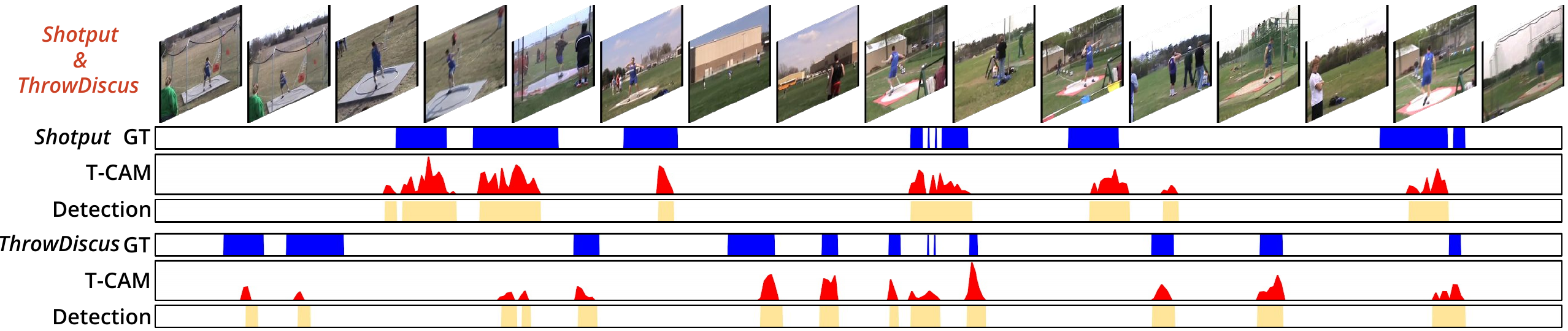} \\
    \includegraphics[width=1\linewidth,height=0.15\linewidth]{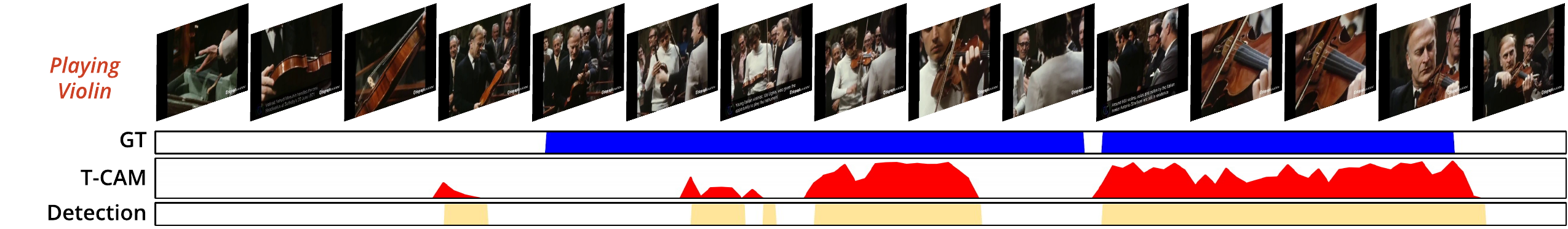} \\
    \includegraphics[width=1\linewidth,height=0.15\linewidth]{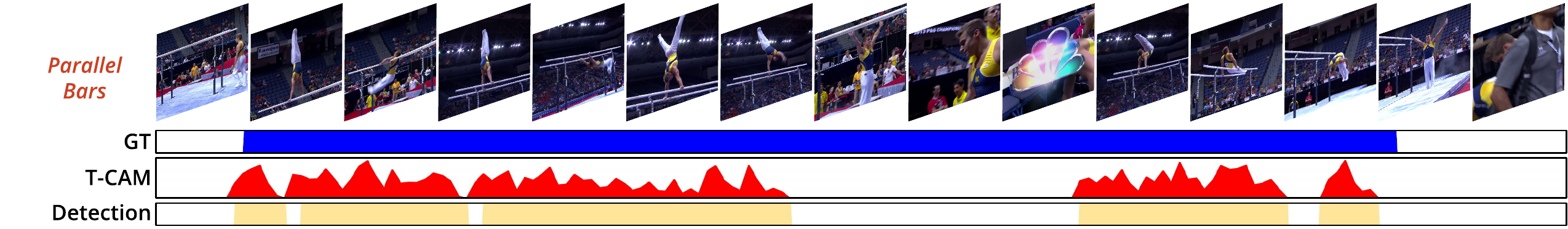}
    \caption{Qualitative temporal action localization results of our 3C-Net approach on example videos from the THUMOS14 and ActivityNet 1.2 datasets. For each video, we show the example frames in the top row, ground-truth segments indicating the action instances as GT and the class-specific confidence scores over time as T-CAM (for brevity, only the thresholded T-CAM is shown). Action segments predicted using the T-CAM are denoted as Detection. Examples show different scenarios: multiple instances of same action (first video), visually-similar multiple action categories (second video) and long duration activities (third and fourth video). Our approach achieves promising localization performance on these variety of actions.}\vspace{-0.3cm}
    \label{fig_qual_res}
\end{figure*}

\subsection{Qualitative Analysis\label{sec_qual_res}}
We now present the qualitative analysis of our 3C-Net approach.
Fig.~\ref{fig_qual_res} shows the qualitative temporal action localization results of our 3C-Net on example videos from the THUMOS14 and ActivityNet 1.2 datasets. For each video, example frames are shown in the top row. GT denotes the ground-truth segments. The category-specific confidence scores over time are indicated by T-CAM. Detection denotes the action segments predicted using the T-CAM.  The top two videos are from  THUMOS14. The multiple instances of \emph{HighJump} action (first video) are accurately localized by our 3C-Net. The second video contains visually similar multiple actions (\emph{Shotput} and \emph{ThrowDiscus}) and has overlapping ground-truth annotations. In this case, 3C-Net mostly localizes the two actions accurately. 

The bottom two examples from the ActivityNet 1.2 dataset contain long duration activities from \emph{Playing Violin} and \emph{Parallel Bars} categories. Observing the T-CAM progression in both videos, we see that the proposed framework detects the action instances reasonably well. For \emph{Playing Violin} video, prediction with respect to the second instance is correctly detected, while the first instance is partially detected. This is due to imprecise annotation of the first instance which has some segments without the playing activity. In \emph{Parallel Bars} video, a single action instance is annotated. However, the video contains an activity instance followed by background segments without any action and ends with the replay of the first action instance. This progression of \emph{activity-background-activity} has been clearly identified by our approach as observed in the T-CAM. These results suggest the effectiveness of our approach for the problem of temporal action localization. We observe common failure reasons to be extreme scale change, visually similar actions confusion and temporally quantized segments for I3D feature generation. Few failure instances in Fig.~\ref{fig_qual_res} are: detections having minimal overlap with the GT (first two detected instances of \emph{ThrowDiscus}), false detections (third and fourth detected instances of \emph{ThrowDiscus}) and multiple detections (first two detected instances of \emph{Parallel Bars}).

\section{Conclusion}
We proposed a novel formulation with classification loss, center loss and counting loss terms for weakly-supervised action localization. We first proposed to use a class-specific attention-based feature
aggregation strategy to utilize multi-label videos for
training with center loss. We further introduced a counting loss term to leverage video-level action count information. To the best of our knowledge, we are the first to propose a formulation with multi-label center loss and action counting loss terms for weakly-supervised action localization. Experiments on two challenging datasets clearly demonstrate the effectiveness of our approach for both action localization and classification.

{\small
\bibliographystyle{ieee_fullname}
\bibliography{egbib}
}

\end{document}